\documentclass[10pt,twocolumn,letterpaper]{article}

\usepackage{iccv}

\usepackage{times}
\usepackage{epsfig}
\usepackage{graphicx}
\usepackage{amsmath}
\usepackage{amssymb}
\usepackage{float}
\usepackage{wrapfig,lipsum,booktabs}
\usepackage{multirow}
\usepackage{color} 
\usepackage{comment}
\usepackage{makecell}
\usepackage{float}
\usepackage{xcolor}

\def\eg{\textit{e.g.}}
\def\ie{\textit{i.e.}}

\def\etal{\textit{et al.}}

\usepackage[pagebackref=true,breaklinks=true,letterpaper=true,colorlinks,bookmarks=false]{hyperref}

\iccvfinalcopy 

\def\iccvPaperID{8} 

\ificcvfinal\fi

\begin{document}

\title{A Unified Efficient Pyramid Transformer for Semantic Segmentation}

\author{Fangrui Zhu\thanks{Work done during an internship at Amazon.}~\textsuperscript{\rm 1}, Yi Zhu\textsuperscript{\rm 2}, Li Zhang\textsuperscript{\rm 1}, Chongruo Wu\textsuperscript{\rm 3}, Yanwei Fu\textsuperscript{\rm 1}, Mu Li\textsuperscript{\rm 2} \\
\textsuperscript{\rm 1} School of Data Science, Fudan University\\
\textsuperscript{\rm 2} Amazon Web Services \\
\textsuperscript{\rm 3} University of California, Davis\\
}

\maketitle


\begin{abstract}
Semantic segmentation is a challenging problem due to difficulties in modeling context in complex scenes and class confusions along boundaries. 
Most literature either focuses on context modeling or boundary refinement, which is less generalizable in open-world scenarios. In this work, we advocate a unified framework~(UN-EPT) to segment objects by considering both context information and boundary artifacts. We first adapt a sparse sampling strategy to incorporate the transformer-based attention mechanism for efficient context modeling. In addition, a separate spatial branch is introduced to capture image details for boundary refinement. The whole model can be trained in an end-to-end manner. We demonstrate promising performance on three popular benchmarks for semantic segmentation with low memory footprint. Code will be released soon.

\end{abstract}

\section{Introduction}
\label{sec:introduction}

\begin{figure}[t]
\begin{center}
   \includegraphics[width=0.45\textwidth]{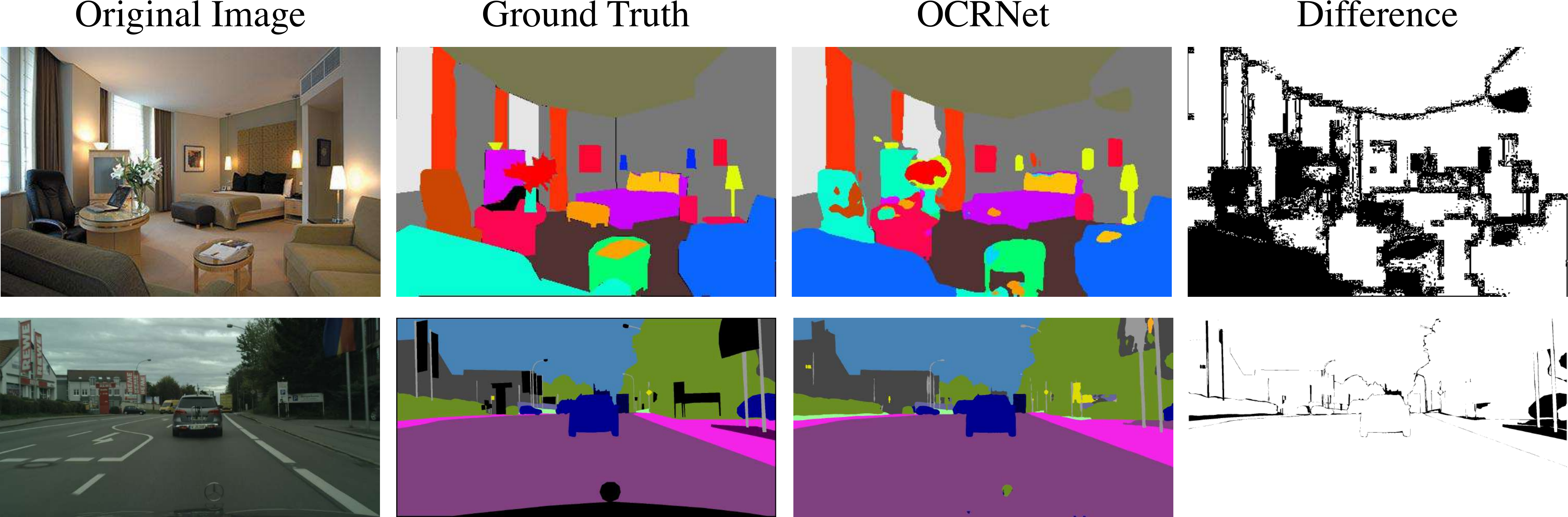}
\end{center}
\vspace{-10pt}
   \caption{
   Comparisons of the predictions from OCRNet and the ground truth, and visualizations of their differences. The sample from ADE20K (top) suffers mainly from class confusion, while most errors from Cityscapes~(bottom) are along the boundary.
   } 
\label{fig:intro}
\end{figure}

Semantic segmentation is the task of dense per-pixel predictions of semantic labels.
Starting from Fully Convolutional Network~(FCN)~\cite{Long2015FCN}, there has been significant progress in model development~\cite{zhao2017pyramid,chen2018encoder,li_EMA19,fu2019dual,yuan2019object,zhu2020csst}. 
However, different datasets  are in favour of distinctive methods. 
Taking Fig.~\ref{fig:intro} as an example, datasets like ADE20K~\cite{mottaghi2014role,zhou2017scene} with a larger number of classes often encounter the problem of confusion among similar object classes, and thus require context modeling and global reasoning~\cite{yu2020context}. 
On the contrary, datasets like Cityscapes~\cite{cordts2016cityscapes} with a less number of classes but  high image  resolution often  suffer from boundary ambiguity,  demanding careful handling in object boundaries~\cite{yuan2020segfix,zhu2019improving,xiangtl_decouple}. 
Hence, this leads to two separate lines of work in this area, \ie,~context modeling~\cite{yuan2018ocnet,zhang2018context,zhao2018psanet,fu2019dual,huang2019ccnet,zhang2019co,yu2020context} and exploiting boundary information~\cite{bertasius2016semantic,chen2016semantic,ruan2019devil,takikawa2019gated}\footnote{Note that these two lines of research are not exclusive, and can be complementary most of the time.}.
It is thus desirable to have a single method that can jointly optimize them. 
In this work, we introduce such a unified framework with a two-branch design: context branch and spatial branch.




Motivated by the strong capability in capturing long-range contextual dependencies, we extend transformers ~\cite{vaswani2017attention} to model the per-pixel classification task as a set prediction problem. That is, translating a sequence of RGB pixels into a sequence of object class labels, representing the segmentation mask. 
Despite reasonable and promising, it is nontrivial to directly apply transformers to such a task.
Particularly, as a dense prediction task, semantic segmentation requires:~(1)~large resolution input image, and (2)~the ability of reasoning both local details and the global scene. However, flattening the raw RGB image or even a smaller feature map will result in a much longer sequence than the ordinary linguistic sentence. Fitting such a long sequence for transformer demands prohibitive GPU memory footprint~\cite{zheng2020rethinking}.
Furthermore, in order to capture local image details, we need to provide input-adaptive context queries to the transformer decoder.

To address the first issue, we introduce an efficient transformer-based module for semantic segmentation. Specifically, to reduce the memory footprint,  we adapt a sparse sampling strategy~\cite{zhu2020deformable} to enforce each element in a sequence only attending to a small set of elements. The intuition is that only informative surrounding pixels are needed to classify a specific pixel. With this memory-efficient attention module, we can also bring pyramid feature maps from different stages of the backbone to enrich multi-scale information. We term this structure as Efficient Pyramid Transformer~(EPT). 
To address the second issue, we introduce a lightweight spatial branch to capture local image details. Specifically, our spatial branch consists of a three-layer backbone and two heads. We use features from the backbone as context queries to enforce the decoder input adaptive.
The two heads are optimized to localize the boundary pixels and eventually used to refine our initial segmentation results from the context branch.

At this point, we present UN-EPT, a UNified Efficient Pyramid Transformer network, to improve both context modeling and boundary handling for semantic segmentation.
With UN-EPT, we outperform previous literature with the same backbone~(\ie,~ResNet50) on ADE20K by a large margin. When combined with a stronger backbone DeiT~\cite{touvron2020training}, our method can achieve an mIoU of $50.5$.
Notably, our model requires neither pretraining on  large-scale dataset~(\eg,~ImageNet22K), nor the huge memory cost (\eg,~our best model only consumes 8.5G memory during training). 
This clearly demonstrates the effectiveness of our approach.
Our contributions can be summarized as: 
\begin{itemize}
    \item We adapt an attention module, termed efficient pyramid transformer, to fully exploit context modeling for semantic segmentation.
    \item We introduce a spatial branch to provide input-adaptive information and refine object boundaries for final segmentation mask prediction. 
    \item We present a unified framework for both context modeling and boundary handling, which achieves promising results on three benchmark datasets: ADE20K, Cityscapes and PASCAL-Context.
\end{itemize}

\section{Related work}
\label{sec:related_work}

\noindent \textbf{Context modeling}
Starting from the seminal work of FCN~\cite{Long2015FCN}, there has been significant progress in models for semantic segmentation. In order to explore context dependencies for improved scene understanding, recent works have focused on exploiting object context by pyramid pooling~\cite{zhao2017pyramid,chen2017deeplab,chen2017rethinking,chen2018encoder,Yang2018DenseASPP}, global pooling~\cite{zhang2018context,yu2018bisenet,yu2018learning,Liu2017parsenet} and attention mechanism~\cite{zhang2020resnest}. In terms of the attention mechanism, earlier works~\cite{zhang2018context,yu2018bisenet} adopt channel attention similar to SENet~\cite{hu2018squeeze} to reweight feature maps as well as learning spatial attention~\cite{yuan2018ocnet,zhao2018psanet,zhang2019co,yu2020context}.
Most of them directly learn the attention on top of the last convolutional features for context modeling due to GPU memory constraints.
In this paper, we exploit a sparse sampling strategy to alleviate the computational cost of attention so that we can fit a long sequence into a transformer.
\cite{zhang2020dynamic,yu2020representative} are two closely related works, also introducing efficient sampling method for segmentation. However, both of them still conduct the sampling in a single scale layer while we adopt sampling across multiple pyramid layers for effective context modeling.

\begin{figure*}
\begin{center}
\includegraphics[width=0.95\textwidth]{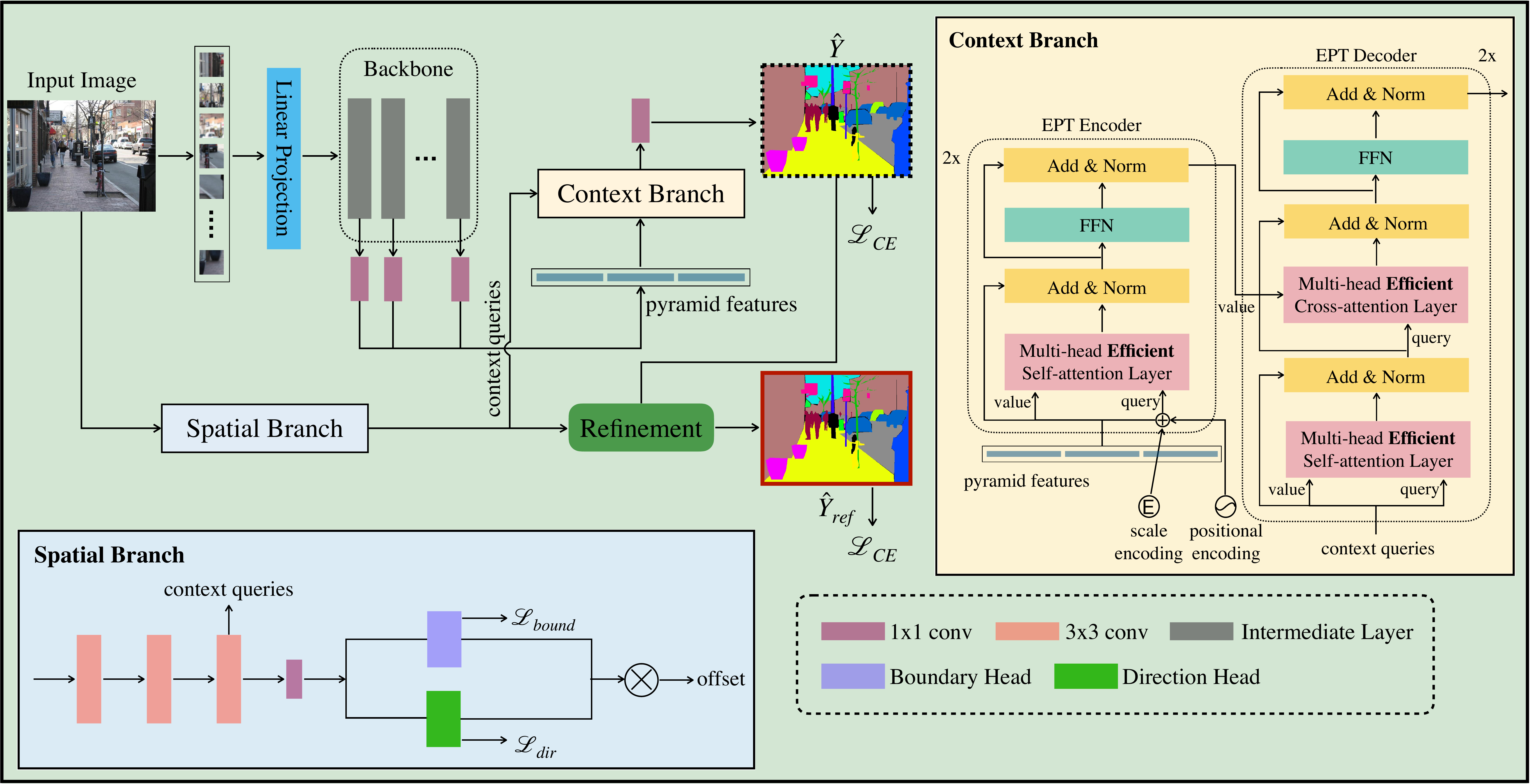}
\end{center}
\vspace{-5pt}
   \caption{UN-EPT architecture learned in an end-to-end manner. 
   Different backbones can be utilized to extract pyramid visual features~(\eg,~ResNet, transformer). We show DeiT here for illustration.
  Image features are flattened into a long sequence and fed into the context branch to 
  obtain the initial segmentation result $\hat{Y}$.
  The spatial branch extracts features for dynamic context queries and the boundary information to refine $\hat{Y}$. The final refined segmentation output is $\hat{Y}_{ref}$.
  }
\label{fig:architecture}
\end{figure*}

\noindent \textbf{Boundary handling}
Previous works focused on either localizing semantic boundaries~\cite{liu2017richer,yu2017casenet,yu2018simultaneous,acuna2019devil} or refining boundary segmentation results~\cite{bertasius2016semantic,takikawa2019gated,ding2019boundary,ruan2019devil}. They are often designed for images with high resolution,
while less useful for modeling contexts, which are prone to error with numerous classes existing. 
Extensive studies~\cite{li2016iterative,gidaris2017detect,kuo2019shapemask,yuan2020segfix} have proposed refinement mechanism to obtain fine segmentation maps from coarse ones, but most of them depend on particular segmentation models. 
\cite{yuan2020segfix} proposes a model-agnostic segmentation refinement mechanism that can be  applied to any approaches. However, it still needs re-training and remains a post-processing method. 
Inspired by \cite{yuan2020segfix}, we propose a dedicated spatial branch to capture more image details, so that we provide dynamic context queries for the decoder input and utilize boundary information for refinement. Importantly, our method is end-to-end trainable, and can handle the case of a large number of object categories and high resolution images, simultaneously.  

\noindent \textbf{Transformer}
Transformer is a powerful model for capturing long-range contextual dependencies, which is firstly introduced in~\cite{vaswani2017attention} for machine translation. After that, it has been widely adopted and becomes the de-facto standard in natural language processing~\cite{devlin2018bert,radford2018improving,radford2019language,brown2020language}. 
Recently, researchers start to apply transformer in computer vision~\cite{zhu2020deformable,dosovitskiy2020image,zheng2020rethinking,xie2021segformer,liu2021swin,cheng2021per,wang2021pyramid}.
SETR~\cite{zheng2020rethinking} directly adopts the ViT~\cite{dosovitskiy2020image} model for semantic segmentation.
The recent PVT~\cite{wang2021pyramid}
proposes a versatile transformer backbone suitable for several vision tasks. 
However, the vanilla  transformer is computationally heavy and memory consuming when the sequence length is long, which is not scalable for semantic segmentation. 
The con-current works SegFormer~\cite{xie2021segformer} and Swin Transformer~\cite{liu2021swin} use MLP decoders and shifted windows to improve the model efficiency. 
Different from them, we change the computation strategy of self- and cross-attention and our model jointly considers modeling boundary information, which is specifically suitable for segmentation.
MaskFormer~\cite{cheng2021per} solves the semantic- and instance-segmentation in a unified manner by introducing mask predictions.
Inspired by the rapid progress of efficient transformers~\cite{beltagy2020longformer,kitaev2020reformer,child2019generating,zhu2020deformable}, we propose a unified efficient pyramid transformer model for semantic segmentation in this work.

\section{Method}
\label{sec:method}

This section introduces the proposed UN-EPT network as illustrated in Fig.~\ref{fig:architecture}. Particularly, we present an efficient transformer for modeling contexts in Sec.~\ref{subsec:context}, including an intuitive sparse sampling strategy to compute attention~(Sec.~\ref{sec:sparse_sampling}). 
We then apply pyramid features naturally to fully explore long-range spatial contexts (Sec.~\ref{sec:multi_scale}). Finally, a spatial branch for segmentation is proposed in Sec.~\ref{subsec:spatial_path} by using  dynamic context queries and boundary refinement.



\subsection{\label{subsec:context}Efficient Transformer for modeling contexts}
\noindent \textbf{Revisiting Transformer}
Transformer takes stacks of self-attention layers in both encoder and decoder.
Positional encoding and multi-head structure are designed for providing position information and modeling relations in a higher dimension.
The standard multi-head attention~\cite{vaswani2017attention} projects the same input feature sequence into different feature spaces: \textit{key}, \textit{query}, \textit{value}, denoted as $K\in \mathbb{R}^{n\times d_{m}}$, $Q\in \mathbb{R}^{n\times d_{m}}$, $V\in \mathbb{R}^{n\times d_{m}}$, where $n$ represents the sequence length and $d_{m}$ is the feature dimension. 
The \textit{attention weights} is computed based on key and query,
\begin{equation}
\label{Eq:weights}
    A_{m}=\textup{softmax}\left ( \frac{Q W_{m}^{Q}\left (KW_{m}^{K} \right)^{\textup{T}}}{\sqrt{d_{model}}} \right )
\end{equation}
where $W_{m}^{Q}, W_{m}^{K}\in \mathbb{R}^{d_{m}\times d_{k}}$ and $A_{m}\in \mathbb{R}^{n\times n}$ denotes the attention weights for head $m=\left \{ 1,2,...,M \right \}$. 
Then we compute the \textit{attention} with weights and value, $ \textup{Attn}_{m}=A_{m} V W_{m}^{V}$.
Here, $W_{m}^{V}\in \mathbb{R}^{d_{m}\times d_{v}}$ and $\textup{Attn}_{m}\in \mathbb{R}^{n\times d_{v}}$ is the attention value of head $m$. Finally, we concatenate the result of each head to obtain \textit{multi-head attention}: $ \textup{MH-Attn}=\left [ \textup{Attn}_{1},...,\textup{Attn}_{M} \right ]W^{O}$.
Here $W^{O}\in \mathbb{R}^{Md_{v}\times d_{m}}$ denotes  linear projection and $\left [\cdot \right ]$ denotes the concatenation operation. 
With the help of multi-head self attention,
the transformer encodes the input feature sequence by letting them attend to each other, where the output feature 
captures long-range contexts, motivating us to apply it to dense per-pixel classification task.

\subsubsection{\label{sec:sparse_sampling}Transformer with sparse sampling}
To model the pixel-to-pixel correlation, transformer brings huge cost in memory space and computation resource when an input sequence is relatively long.
In the case of segmentation, several problems make the transformer inefficient and impractical in real scenarios: (1) Large input size results in a \textit{long} pixel feature sequence, making it impossible to fit in a general GPU. (2) Attending to \textit{all} pixels is unwise and may cause \textit{confusion}. 
For instance, to segment different instances of the same category, each pixel only needs to attend to the instance region which it belongs to, where features of unrelated instances are redundant.

To tackle above issues, we adapt a sparse sampling strategy in~\cite{zhu2020deformable} to semantic segmentation. That is, to force each query pixel to attend only a small set of informative pixels for computing attention.
Give an input image, we pass it into a backbone network~(\eg,~ResNet50~\cite{he2016deep} or DeiT~\cite{touvron2020training}) for a feature map $I\in \mathbb{R}^{H\times W\times C}$.
Then $I$ is passed by a $1\times 1 ~\textup{conv}$ to reduce channel dimension and sent into the transformer.
The input is a sequence of flattened pixel features, denoted as $X\in \mathbb{R}^{HW\times d_{model}}$. 
We map $X$ to query $Q\in \mathbb{R}^{HW\times d_{model}}$ and value $V\in \mathbb{R}^{HW\times d_{model}}$ features with two parameterized matrice, respectively. Different from the standard self-attention, we do not need key features to compute attention weights here. Instead, our attention weights
are learned from query features by a projection matrix. 

Then, for each query pixel~$q \in Q$ with value as $v$, its attention of head $m$ is computed by
\begin{equation}
\label{Eq:ss_weights}
    \textup{Attn}_{mq}= \textup{softmax} \left (
    \sum_{n=1}^{N} w_{nq} \right ) v_{\left \langle  c_{q}+\Delta _{n} \right \rangle}
\end{equation}
where $N$ is the number of sparse sampling pixels and $c_{q}$ is the coordinates of $q$. $\left \langle \cdot  \right \rangle$ and $\Delta _{n}$ denote the interpolation method and sampling offsets. 
$w_{nq}$ is the attention weights for $q$ and the $n$-th sampled key element. For simplicity, we omit the linear projection matrice in Eq.~\ref{Eq:weights}. 
Then, for each query element in $Q$, it attends to $N$ value features for calculating attention rather than $HW$ features in $V$, reducing the computation complexity from $O(n^{2})$ to $O(kn)$~($n=HW$).

To further reduce computation cost, 
the $\Delta _{n}$ and attention weights $w_{nq}$~$(n=1,2, ...,N)$ for $N$ value features are mapped directly from $Q$, which can be written as
\begin{equation}
\label{Eq:deform_weights}
    A_{m}=\textup{softmax}\left ( Q W_{m}^{Q}U_{m}^{wts} \right ),
\end{equation}

\begin{equation}
\label{Eq:deform_attn}
    \textup{Attn}_{m}=A_{m} \left ( V W_{m}^{V} \right )_{\left \langle Q W_{m}^{Q}U_{m}^{pos} \right \rangle}
\end{equation}
where $U_{m}^{wts}\in \mathbb{R}^{d_{k}\times N}$ projects the query feature into attention weights and $U_{m}^{pos}\in \mathbb{R}^{d_{k}\times 2N}$ denotes relative positions of key-query in $x$-axis and $y$-axis, as illustrated in \cite{zhu2019empirical}.

In addition, to inform the model with image spatial information, we add sine and cosine positional encodings on the query $Q$.
In practice, the transformer encoder uses the stack of multi-head self-attention layers and feed forward layers to encode the input feature map $X$, and obtains $X_{enc}$ of the same size.
On the decoder side, we take a feature $C\in \mathbb{R}^{HW\times d_{model}}$ as input, serving as context queries. It firstly performs self-attention as in Eq.~\ref{Eq:deform_attn} and computes cross-attention with the encoder output $X_{enc}$ to produce final results $\hat{Y}$. The whole process can be written as
\begin{equation}
    \hat{Y} = \Phi _{dec}\left ( \Phi _{enc}\left ( X\right ), C \right )
\end{equation}
where $\hat{Y}\in \mathbb{R}^{HW\times d_{model}}$ is then up-sampled and computed cross entropy loss with the ground truth.
The computation manner of cross attention is the same with self-attention, where the output of decoder self-attention serves as query and $X_{enc}$ is the value feature. 

\subsubsection{\label{sec:multi_scale}Efficient pyramid transformer}
Due to the memory constraint, seldom approaches use multi-scale information for segmentation problem.
Here, our model allows us to incorporate pyramid features naturally.
As shown in Fig.~\ref{fig:architecture}, $1 \times 1$ convolutional layers are adopted separately to obtain pyramid feature maps of the same channel size from the backbone, 
denoted as $\left \{ X^{l} \right \}_{l=1}^{L}$, where $X^{l}\in \mathbb{R}^{H_{l}W_{l}\times C}$. 
To feed into the transformer encoder, they are concatenated in a long sequence $X_{ms} \in \mathbb{R}^{L_{ms} \times d_{model}}$. 
Similarly, we use linear projections to map $X_{ms}$ to query $Q$ and value $V$ features, respectively. 
We add positional encoding and scale encoding on $Q$, providing model with more spatial information, as shown in the context branch of Fig.~\ref{fig:architecture}. The scale encoding is a learnt embedding of size $L \times d_{model}$.
For each query pixel $q$, similar with Eq.~\ref{Eq:ss_weights}, it attends to a set of pixels on the feature map of each scale, where the attention is computed as 

\begin{equation}
\label{Eq:ms_weights}
    \textup{Attn}_{mq}=\textup{softmax} \left(
    \sum_{l=1}^{L} \sum_{n=1}^{N} w_{lnq} \right ) v_{\left \langle  c_{q}+\Delta _{ln} \right \rangle} 
\end{equation}
where $w_{lnq}$ denotes the attention weight of $q$ and $n$-th sampled key element on the scale $l$.
$\Delta _{ln}$ denotes sampling offsets of pixels from each feature scale.
Thus, $q$ attends to $N \times L$ pixels, still much less than attending to all $HW$ pixel features. 
In practice, we set $U_{m}^{wts}\in \mathbb{R}^{d_{k}\times NL}$ and $U_{m}^{pos}\in \mathbb{R}^{d_{k}\times 2NL}$ in Eq.~\ref{Eq:deform_weights} and \ref{Eq:deform_attn}.

The encoder output $X_{enc}\in \mathbb{R}^{L_{ms} \times d_{model}}$ is then fed into the decoder to compute cross attention. We keep the decoder input~(context queries) of size $H_{L}W_{L} \times d_{model}$ to recover the image resolution through computing cross attention with $X_{enc}$.
Here, context queries reason from pyramid features and select informative pixels to generate predictions. The output is up-sampled to compute the loss. With pyramid features, we are able to model stronger spatial contexts.

\subsection{\label{subsec:spatial_path} Dynamic learnable spatial branch}
Compared to image classification and object detection, one major difference of using transformer for semantic segmentation is that we need high frequency image details for the dense prediction.  
However, when the transformer is applied to vision tasks, like DETR~\cite{carion2020end}, input queries for decoders are fixed embeddings for all images, which are designed to learn the global information for the dataset. This specific information may not be suitable for segmentation problem. To alleviate it, we introduce a spatial branch to adapt to various input images.
We further leverage the spatial branch to capture the boundary information, since its outputs are in relatively high resolutions and maintain more image details. We adopt two additional heads to predict boundary pixels~(boundary head) and the corresponding interior pixel for each boundary pixel~(direction head). The offset generated from these two predictions is used to refine the segmentation result from the context branch.

\noindent \textbf{Context queries}
The decoder input serves as context queries, which has different design choices.
The naive choice~\cite{carion2020end} is using a random initialized embedding.
 However, that embedding will be the same for all input images at inference time, which cannot provide sufficient contextual information. Thus, we introduce dynamic contexts as the decoder input to flourish the self- and cross-attention.
As shown in the spatial branch of Fig.~\ref{fig:architecture}, our spatial branch contains three $3 \times 3$ convolutional layers, followed by batch normalization and ReLU, to extract representations from input images.
This branch produces output feature map that is 1/8 of the original image, encoding rich and detailed spatial information due to the large spatial size.
Empirically, we choose the intermediate feature from spatial branch to be context queries.

\noindent  \textbf{Boundary refinement}
Inspired by the success of SegFix~\cite{yuan2020segfix}, we adopt a boundary head and a direction head to extract boundary information from the output feature map of the backbone. Similarly, the boundary head contains $1 \times 1 ~\textup{Conv} \rightarrow \textup{BN} \rightarrow \textup{ReLU}$ with 256 output channels. A linear classifier~($1 \times 1 ~\textup{Conv}$) and up-sampling are further applied to generate the final boundary map of size $H \times W \times 1$. The boundary loss is the binary cross-entropy loss, denoted as $\mathcal{L}_{bound}$. For the direction head, we directly discrete directions by dividing the entire direction range to $m$ partitions as the same as the ground truth~($m=8$ by default). The direction head contains $1 \times 1 ~\textup{Conv} \rightarrow \textup{BN} \rightarrow \textup{ReLU}$ with 256 output channels. A linear classifier ~($1 \times 1 ~\textup{Conv}$) and up-sampling are further applied to generate the final direction map of size $H \times W \times m$. The direction map is multiplied by the boundary map to ensure direction loss is only applied on boundary pixels. The cross-entropy loss is used to supervise the discrete directions, denoted as direction loss $\mathcal{L}_{dir}$. For the refinement process, we convert the predicted direction map to the offset map of size $H \times W \times 2$. The mapping scheme follows~\cite{yuan2020segfix}. Then, we generate the refined label map through shifting the coarse label map with the grid-sample scheme~\cite{jaderberg2015spatial}. There are multiple mechanisms to generate ground truth for the boundary maps and the direction maps. In this work, we mainly use the conventional distance transform~\cite{kimmel1996sub}, similar with \cite{yuan2020segfix}. The final loss function is
\begin{equation}
    \mathcal{L} = \lambda_{1} \mathcal{L}_{CE}(Y, \hat{Y}) + \lambda_{2} \mathcal{L}_{CE}(Y, \hat{Y}_{ref}) + \lambda_{3} \mathcal{L}_{bound} + \lambda_{4} \mathcal{L}_{dir}
\end{equation}
where $\hat{Y}_{ref}$ is the refined segmentation result. In practice, we set $\lambda_{1}=1$, $\lambda_{2}=1.5$, $\lambda_{3}=3$, $\lambda_{4}=0.7$.


\section{Experiments}
We first introduce experimental settings including datasets in Sec.~\ref{exp:data} and implementation details in Sec.~\ref{exp:imple}. We then analyze our results on three benchmark datasets,
by comparing with state-of-the-art methods in Sec.~\ref{exp:analysis_result}. 
Following that, we conduct several ablation studies on ADE20K in Sec.~\ref{exp:ablation}.

\subsection{\label{exp:data} Datasets}
Our proposed UN-EPT is evaluated on three standard segmentation benchmarks, \ie,~ADE20K dataset~\cite{zhou2017scene}, PASCAL-Context dataset~\cite{mottaghi2014role} and Cityscapes dataset~\cite{cordts2016cityscapes}.

\noindent
\textbf{ADE20K dataset~\cite{zhou2017scene}}
 is a recent scene parsing benchmark containing dense labels of 150 object category labels. It is challenging due to its large number of classes and existence of multiple small objects in complex scenes.
 This dataset includes 20K/2K/3K images for training, validation and test. We train our model on the training set and evaluate on the validation set.
 
 \noindent
 \textbf{PASCAL-Context dataset~\cite{mottaghi2014role}}
 provides dense annotations for the whole scene in PASCAL VOC 2010, which contains 4998/5105/9637 images for training, validation and test. Following previous works~\cite{zhang2018context,zhu2019asymmetric}, we use 60 class labels~(59 object categories plus background) for training and testing. Our results are reported on the validation set.
 
 \noindent
 \textbf{Cityscapes dataset~\cite{cordts2016cityscapes}}
 is a large urban street dataset particularly created for
scene parsing, including 19 object categories. It contains 2975/500 fine annotated images for training and validation. Additionally, it has 20000 coarsely annotated training images, but we only use fine annotated training images and report results on the validation set.
And there are 1,525 images for testing. 
We only use fine annotated training images in our setting and we report the results on test set which contains 1525 images.

\begin{table}[t]
\scriptsize
\centering
\begin{tabular}{l|c|c|c|c}
    \toprule
    Method & Reference & Backbone & mIoU & pixAcc\\
     \midrule
     PSPNet~\cite{zhao2017pyramid} & CVPR2017 & ResNet50 & 41.7 & 80.0 \\
    PSANet~\cite{zhao2018psanet}  & ECCV2018 & ResNet50 & 42.9 & 80.9  \\
    UperNet~\cite{xiao2018unified}  & ECCV2018 & ResNet50  & 41.2 & 79.9 \\
    EncNet~\cite{zhang2018context} & CVPR2018 & ResNet50 & 41.1 & 79.7\\
    CFNet~\cite{zhang2019co} & CVPR2019 & ResNet50  & 42.9 & -  \\
    CPNet~\cite{yu2020context}   & CVPR2020 & ResNet50 & 44.5 & 81.4   \\
    \midrule
    Ours & - & ResNet-50 & \textbf{46.1}  & \textbf{81.7} \\
    \midrule
   RefineNet~\cite{lin2017refinenet} & CVPR2017 & ResNet101 & 40.2 & -\\
    PSPNet~\cite{zhao2017pyramid} & CVPR2017 & ResNet101 & 43.3 & 81.4 \\
    SAC~\cite{zhang2017scale} & ICCV2017 & ResNet101 & 44.3 & 81.9    \\
    UperNet~\cite{xiao2018unified}  & ECCV2018 & ResNet101  & 42.7 & 81.0 \\
    DSSPN~\cite{liang2018dynamic} & CVPR2018 & ResNet101 & 43.7 & 81.1   \\
    PSANet~\cite{zhao2018psanet}  & ECCV2018 & ResNet101 & 43.8 & 81.5  \\
    EncNet~\cite{zhang2018context} & CVPR2018 & ResNet101 & 44.7 & 81.7\\
    ANL~\cite{zhu2019asymmetric} & ICCV2019 & ResNet101 & 45.2 & -   \\
    CCNet~\cite{huang2019ccnet} & ICCV2019 & ResNet101 & 45.2 & - \\
    CFNet~\cite{zhang2019co} & CVPR2019 & ResNet101  & 44.9 & -  \\
    CPNet~\cite{yu2020context}   & CVPR2020 & ResNet101 & 46.3 & 81.9   \\
    OCRNet~\cite{yuan2019object} & ECCV2020 & ResNet101 & 45.3 & -   \\
    Efficient FCN~\cite{liu2020efficientfcn} & ECCV2020 & ResNet101 & 45.3 & -  \\
    ResNeSt~\cite{zhang2020resnest} & arXiv2020 & ResNeSt200 & 48.4 & - \\
    SETR~\cite{zheng2020rethinking} & CVPR2021 & T-large & 50.2 & 83.5\\
    \midrule
    Ours & - & DeiT  & \textbf{50.5}& \textbf{83.6}\\
    \bottomrule
    \end{tabular}
    \vspace{6pt}
    \caption{Quantitative evaluations on the ADE20K validation set. }
    \label{tab:quan_ade}
    \vspace{-10pt}
\end{table}

\subsection{\label{exp:imple} Implementation details}
We implement our experiments with PyTorch~\cite{zhu2019asymmetric} and MMSegmentation~\cite{mmseg_tool} open source toolbox.

\begin{figure*}
    \centering
    \includegraphics[scale=0.27]{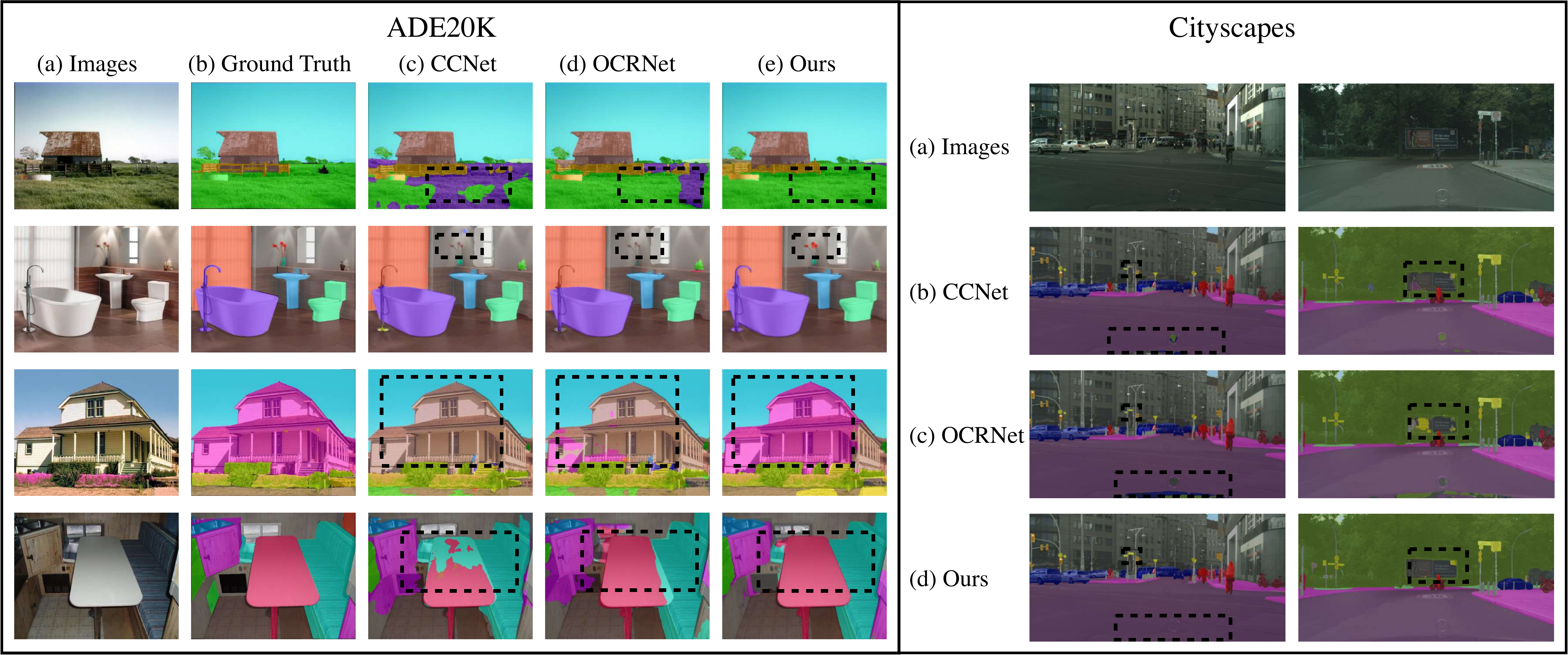}
    \caption{Qualitative results of  proposed UN-EPT on three datasets.
    We compare our method with CCNet~\cite{huang2019ccnet} and OCRNet~\cite{yuan2019object}. We visualize them with models released from MMSegmentation.
    On ADE20K and PASCAL-Context, our model deals with contexts and small objects better. And our model handles the boundary artifacts better on Cityscapes. See black dashed boxes for differences, best zoomed in.}
    \label{fig:qual_vis}
        \vspace{-0.15in}
\end{figure*}

\noindent \textbf{Network structure}
For the ResNet50~\cite{he2016deep} backbone, we use ImageNet pre-trained weights as initialization and apply the dilation strategy~\cite{chen2014semantic,yu2015multi} to obtain 1/8 output size of the input image. The $1 \times 1$ convolutional layer is used to reduce the channel dimension of backbone features to $d_{model}=256$. We take intermediate features from stage 3 to 5 in the backbone to obtain pyramid features. 
For the stronger backbone DeiT~\cite{touvron2020training}, we use the base version with a 12-layer encoder. We set the patch size as $16 \times 16$ and adapt the positional encoding with bilinear interpolation. For pyramid features, we extract pyramid features from outputs of the 4th, 8th, 12th layer followed by $1 \times 1$ convolutional layers. The embedded dimension of the encoder layer is 768, with the head number as 12. We use the model weights from \cite{touvron2020training} as pretrained weights, which are trained with ImageNet1K. 
The dimensions of positional encoding and scale encoding are the same with $d_{model}$ and they are added with pyramid features to serve as transformer encoder inputs.
For EPT, we use 2 encoders and decoders with head number being $M=8$ and $d_{k}=d_{v}=32$.
The hidden dimension of the feed forward layer is 2048. We model the residual inside transformer layers and layer normalization is applied.
The dropout strategy is applied after the linear layer in the attention computation and after two linear layers of the feed forward module. 
For UN-EPT, we empirically set $N=16$ and $L=3$. That is, selecting 16 pixels on the feature map of each scale for computing attention. On the encoder side, we map the query feature~(image feature) to offsets for sampling pixels. On the decoder side, we also map the query feature to offsets of sampling features.
The output predictions are upsampled 8 times by bilinear interpolation for cross entropy loss. The whole framework is trained in an end-to-end manner.

\noindent \textbf{Evaluation metrics}
We use standard segmentation evaluation metrics of pixel accuracy~(pixAcc) and mean Intersection of Union~(mIoU). For the ADE20K dataset, we follow the standard benchmark~\cite{zhou2017scene} to ignore background pixels for computing mIoU. 
For Cityscapes, we report our results by submitting to the test server.

\noindent \textbf{Training}
For data augmentation, we randomly scale the image with ratio range from 0.5 to 2.0. Random horizontal flipping, photometric distortion and normalization are further adopted to avoid overfitting.
Images are then cropped or padded to the same size to feed into the network~($480 \times 480$ for ADE20K and PASCAL-Context and $768 \times 768$ for Cityscapes). 
We train 160k iterations for the ADE20K dataset and 80k iterations for the Cityscapes dataset. 
The base learning rate is 1e-4 except that the base learning rate of backbone layers is 1e-5.
The learning rates are dropped at 2/3 iteration with 0.1. 
We use AdamW optimizer with weight decay being 1e-4, $\beta_{1}=0.9$ and $\beta_{2}=0.999$. 
We set batch size as 16 for all experiments and synchronized batch normalization is utilized.
For all ablation experiments, we run the same training recipe five times and report the average mIoU and pixAcc.

\noindent \textbf{Inference}
At inference time, following \cite{zhao2017pyramid,yu2018learning,zhang2018context,yu2020context}, we
apply scaling~(scale ratio: $\left \{ 0.5, 0.75, 1.0, 1.25, 1.5, 1.75 \right \}$), flipping and normalization to augment test inputs and obtain the average predictions as final outputs.

\begin{table}[]
\scriptsize
\centering
\begin{tabular}{l|c|c}
    \toprule
    Method  & Backbone & mIoU \\
    \midrule
    PSPNet~\cite{zhao2017pyramid}  & ResNet101 & 78.5 \\
    DeepLabv3~\cite{chen2017rethinking}~(MS)  & ResNet101 & 79.3 \\
    PointRend~\cite{alex2019pointrend}  & ResNet101 & 78.3 \\
    OCRNet~\cite{yuan2019object}  & ResNet101 & 80.6\\
    Multiscale DEQ~\cite{bai2020multiscale}~(MS)  & MDEQ & 80.3 \\
    CCNet~\cite{huang2019ccnet}  & ResNet101 & 80.2 \\
    GCNet~\cite{cao2019gcnet} & ResNet101 & 78.1 \\
    Axial-DeepLab-XL~\cite{wang2020axial}~(MS) & Axial-ResNet-XL & 81.1 \\
    Axial-DeepLab-L~\cite{wang2020axial}~(MS) & Axial-ResNet-L & 81.5 \\
    SETR~\cite{zheng2020rethinking}~(MS) & T-large & 82.2 \\
    \midrule
    Ours~(80k, MS) & ResNet50 & 79.8 \\
    Ours~(80k, MS)  & DeiT & \textbf{82.9}  \\
    \bottomrule
    \end{tabular}
    \vspace{6pt}
    \caption{Quantitative evaluations on the Cityscapes validation set~(training iterations:~80k, MS: Multi-scale inference).}
    \label{tab:quan_cityscape_val}
\end{table}

\begin{table}[]
\scriptsize
\centering
\begin{tabular}{l|c|c|c}
    \toprule
    Method & Reference & Backbone & mIoU \\
    \midrule
    RefineNet~\cite{lin2017refinenet} & CVPR2017 &ResNet101 & 73.6\\
    PSPNet~\cite{zhao2017pyramid} & CVPR2017 & ResNet101 & 78.4 \\
    SAC~\cite{zhang2017scale} & ICCV2017& ResNet101 & 78.1 \\
    BiSeNet~\cite{yu2018bisenet} & ECCV2018 & ResNet101 & 78.9 \\
    PSANet~\cite{zhao2018psanet}  & ECCV2018 & ResNet101 & 80.1 \\
    ANL~\cite{zhu2019asymmetric} & ICCV2019 & ResNet101 & 81.3 \\
    CPNet~\cite{yu2020context}  & CVPR2020 & ResNet101 & 81.3 \\
    OCRNet~\cite{yuan2019object} & ECCV2020 & ResNet101 & 81.8 \\
    CDGC~\cite{hu2020class} & ECCV2020 & ResNet101 & 82.0 \\
    SETR~\cite{zheng2020rethinking} & CVPR2021 & T-large & 81.6 \\
    \midrule
    Ours & - & ResNet50 & 80.6  \\
    Ours & - & DeiT & \textbf{82.2}  \\
    \bottomrule
    \end{tabular}
    \vspace{6pt}
    \caption{Quantitative evaluations on the Cityscapes test set.}
    \label{tab:quan_cityscape_test}
\end{table}

\subsection{\label{exp:analysis_result} Comparison to state-of-the-art}

\noindent
\textbf{Results on ADE20K dataset.}
Here, we show both quantitative and qualitative results on ADE20K dataset.  
For ADE20K, the complexity is mainly due to large class number and existence of small objects.
We first report the results trained with a ResNet50 backbone. 
From Tab.~\ref{tab:quan_ade}, we achieve $46.1$ in terms of mIoU, $1.6$ higher than previous state-of-the-art CPNet~\cite{yu2020context} trained with ResNet50.
We also want to emphasize that ours with ResNet50 obtain competitive performance compared to previous methods trained with ResNet101.
In addition, we can achieve state-of-the-art performance with a stronger backbone network, \textit{i.e.} DeiT~\cite{touvron2020training}.
Compared to recent transformer-based method SETR~\cite{zheng2020rethinking}, our DeiT-base model pretrained on ImageNet1K even outperforms its ViT-large model pretrained on ImageNet22K. 

Next we show several visualizations in Fig.~\ref{fig:qual_vis}, our method can model contextual information  well under the circumstance of complex classes, indicating transformer based attention is applicable to explore spatial correlations. Compared with recent methods CCNet~\cite{huang2019ccnet} and OCRNet~\cite{yuan2019object}, our method can distinguish objects without confusing with other object categories, as in the 1st, 3rd, 4th rows of Fig.~\ref{fig:qual_vis}. Besides, our method is able to segment small objects, producing fine-grained results~(the 2nd row in Fig.~\ref{fig:qual_vis}).

\noindent
\textbf{Results on Cityscapes dataset.}
Here, we show both quantitative and qualitative results on Cityscapes dataset.  
We adopt the best recipe in  practice.
Tables~\ref{tab:quan_cityscape_val} and \ref{tab:quan_cityscape_test} show the comparative results on the validation and test set of Cityscapes, respectively.
We achieve $82.2$ in terms of mIoU, with a DeiT structure as the backbone. We can see that our model outperforms most of concurrent works. 
We only train the model with the fine annotated images. 


\noindent
\textbf{Results on PASCAL-Context dataset.}
Table~\ref{tab:quan_pascal_cxt} shows segmentation results on PASCAL-Context. 
We achieve $55.2$ in terms of mIoU, outperforming  DANet~\cite{fu2019dual} and CPNet~\cite{yu2020context} in modeling contexts with either attention mechanisms or contextual priors.
This indicates our advantage in modeling contextual information through an efficient Transformer-based module. 
Note that, SETR outperforms us by a small margin, but with a larger network~(ViT-large) and better pretraining~(ImageNet21K). 

\begin{table}[]
\centering
\scriptsize
\begin{tabular}{l|c|c|c}
    \toprule
    Method & Reference & Backbone & mIoU \\
    \midrule
    FCN-8S~\cite{Long2015FCN} & CVPR2015 & VGG16 & 37.8 \\
    BoxSup~\cite{dai2015boxsup} & ICCV2015 & VGG16 & 40.5  \\
    RefineNet~\cite{lin2017refinenet} & CVPR2017 & ResNet152 & 47.3 \\
    PSPNet~\cite{zhao2017pyramid} & CVPR2017 & ResNet101 & 47.8 \\
    EncNet~\cite{zhang2018context} &CVPR2018 & ResNet101 & 51.7 \\
    DANet~\cite{fu2019dual} & CVPR2019 & ResNet101 & 52.6 \\
    ANL~\cite{zhu2019asymmetric} & ICCV2019& ResNet101 & 52.8 \\
    CPNet~\cite{yu2020context}  & CVPR2020& ResNet101 & 53.9 \\
    OCRNet~\cite{yuan2019object} & ECCV2020 & ResNet101 & 54.8 \\
    Efficient FCN~\cite{liu2020efficientfcn} & ECCV2020& ResNet101 & 55.3 \\
    SETR~\cite{zheng2020rethinking} & CVPR2021 & T-large & \textbf{55.8} \\
    \midrule
    Ours & - & ResNet50 & 49.5 \\
    Ours & - & DeiT & 55.2  \\
    \bottomrule
    \end{tabular}
    \vspace{6pt}
    \caption{Quantitative evaluations on the PASCAL-Context validation set.}
    \label{tab:quan_pascal_cxt}
\end{table}

\begin{table}[t]
\scriptsize
\centering
\begin{tabular}{l|c|c}
    \toprule
    Variants  & pixAcc & mIoU \\
    \midrule
    baseline  & 80.3 & 42.3 \\
    +~pyramid features  & 81.1 & 45.0 \\
    +~spatial path  & 81.7  &  46.1  \\
    +~stronger backbone  & \textbf{83.6}  &  \textbf{50.5} \\
    \bottomrule
    \end{tabular}
    \vspace{6pt}
    \caption{Evaluation of different components on ADE20K validation set. }
    \label{tab:diff_comp}
\end{table}

\begin{table}[t]
\scriptsize
\centering
\begin{tabular}{l|c|c|c}
\toprule
Feature scales~(L)  & 
Sampling points~(N) &
pAcc & mIoU \\
\midrule
\multicolumn{1}{c|}{\multirow{3}{*}{1}} & 4 & 77.5 & 37.5 \\
\multicolumn{1}{c|}{} & 16 & 80.3 &  42.5   \\
\multicolumn{1}{c|}{} & 64 & 80.5 & 42.9    \\
\midrule
\multicolumn{1}{c|}{\multirow{3}{*}{3}} & 4 & 80.6 & 44.1 \\
\multicolumn{1}{c|}{} & 16 & \textbf{81.7} & \textbf{46.1}    \\
\multicolumn{1}{c|}{} & 64 & 80.4 & 45.6    \\
  \bottomrule
\end{tabular}
\vspace{6pt}
\caption{Ablation studies on different sampling points~($N$) and different feature scales~($L$) on ADE20K dataset with a ResNet50 backbone. Note that, boundary refinement is not applied here.}
\label{ablation:diff_number}
\end{table}

\subsection{\label{exp:ablation} Ablation studies}
\noindent
\textbf{Different components.}
To examine the effectiveness of different components, we conduct a series of experiments by adding one component at a time, \textit{e.g.} pyramid features, spatial branch for dynamic context and boundary refinement, a stronger attention backbone. 
All models are trained on ADE20K training set and evaluated on the validation set. As shown in Tab.~\ref{tab:diff_comp}, utilizing pyramid features can improve the mIoU from 42.3 to 45.0 on ADE20K. By further adopting a spatial branch to produce the decoder input and explore boundary information, the mIoU can improve with a 1.1 margin, indicating the benefits in learning from abundant contexts and boundary information.
Besides, we show that a stronger attention model is able to extract useful features for better context modeling, improving the mIoU with a large margin~(4.5).

\noindent
\textbf{Different number of sampling points.}
Next, we study the effect of using different number of sampling points. Firstly, we set $N=4, 16, 64$ for the single scale setting, respectively. As shown in Tab.~\ref{ablation:diff_number}, sampling more points to compute attention improves the mIoU to a certain extent, but the number of 16 is sufficient for boosting the mIoU. For pyramid features, we sample 4, 16, 64 points on feature map of each scale. We find that the number of sampling points does not have a large impact on mIoU performance and sampling 16 points on each single scale feature map consistently produces higher performance. Thus, in practice, we set $N=16$ in experiments. Furthermore, incorporating points from pyramid features is always able to improve the performance, showing the effectiveness of the proposed UN-EPT in learning long-range dependencies.

\begin{table}[!htb]
\centering
\scriptsize
\begin{tabular}{ccccc}
\bottomrule
Method & \#Params. & GFLOPs & mIoU \\
\midrule
  PSANet~\cite{zhao2018psanet}     &    73M  & 239.5 & 43.8  \\
  UperNet~\cite{xiao2018unified}     &    86M  & 226.2 & 44.9   \\
  EncNet~\cite{zhang2018context}     &      56M  & 192.2   & 44.7   \\
  CCNet~\cite{huang2019ccnet}     &    69M & 244.6  &  45.2   \\
  PointRend~\cite{alex2019pointrend}     &   48M &  526.5 &  41.6    \\
  DNLNet~\cite{yin2020disentangled}& 70M & 244.1   & 45.9   \\
  OCRNet~\cite{yuan2019object}     &  71M &   144.8   &  44.9   \\
  SETR~\cite{zheng2020rethinking}     &    401M & -     &   50.2  \\
\midrule
  Ours     & 94M     & 99.1  & \textbf{50.5}   \\
\bottomrule
\end{tabular}
\vspace{6pt}
\caption{\label{tab:comp_complex} Comparison of the number of model parameters. We report numbers from models trained on ADE20K. Models are obtained from MMSegmentation~\cite{mmseg_tool} or re-implemented.
 }
\end{table}

\begin{table}[]
\scriptsize
\centering
\begin{tabular}{l|c|c|c}
    \toprule
    Method & Backbone & Mem Cost & mIoU \\
    \midrule
    Sparse Transformer~\cite{child2019generating} & ResNet50 & 18.6G   & 40.3 \\
    Longformer~\cite{beltagy2020longformer} & ResNet50 & 11.3G  &  39.4 \\
    Reformer~\cite{kitaev2020reformer}  & ResNet50 & 15.5G  &  38.2 \\
    \midrule
    CCNet~\cite{huang2019ccnet} & ResNet50 & 9.8G  & 43.1 \\
    ANL~\cite{zhu2019asymmetric} & ResNet50 & 2.0G  & 42.6 \\
    SETR~\cite{zheng2020rethinking} & T-large & 30.0G & 50.2 \\
    \midrule
    Ours & ResNet50 & 7.0G & 46.1   \\
    Ours & DeiT & 8.5G & \textbf{50.5}   \\
    \bottomrule
    \end{tabular}
    \vspace{6pt}
    \caption{Ablation studies on the memory efficiency of UN-EPT. We report results on ADE20K validation set.}
    \label{tab:ablation_efficient}
    \vspace{-10pt}
\end{table}

\noindent
\textbf{Efficient transformer}
We study the effects of efficient transformers in the NLP area, to compare with the proposed UN-EPT. In particular, we adopt Sparse Transformer~\cite{child2019generating}, Reformer~\cite{kitaev2020reformer} and Longformer~\cite{beltagy2020longformer}, respectively. We adopt their transformer structures in our model. Sparse Transformer~\cite{child2019generating} and  Longformer~\cite{beltagy2020longformer} are similar in adopting strided/dilated sliding window to attend to a set of regions instead of full attention, saving memory as well as computation cost. But positions of the attended points are relatively fixed for a particular pixel, which is not suitable for complex scenes generally in the segmentation task. Reformer~\cite{kitaev2020reformer} reduces the complexity by using locality-sensitive hashing and reversible residual layers. However, it cannot remedy the lack of context modeling in our case. Thus, the efficient transformers in NLP area truly reduce the memory cost in our case, but they fail to boost the performance, compared with our UN-EPT, as shown in Tab.~\ref{tab:ablation_efficient}. Our method takes up 
even less GPU memory and improves the mIoU metric with a large margin. We also compare our UN-EPT with efficient structures for segmentation, \ie,~CCNet~\cite{huang2019ccnet}, ANL~\cite{zhu2019asymmetric}.
The results in Tab.~\ref{tab:ablation_efficient} show that our UN-EPT can continuously show competitive performance. In addition, we highlight our efficiency in adopting Transformer-based attention, compared with the recent work SETR~\cite{zheng2020rethinking}. Our method saves the memory cost to a large extent as well as producing better segmentation results. 
We also compare the number of model parameters and GFLOPs in Tab.~\ref{tab:comp_complex}. Notably, our UN-EPT has the lowest GFLOPs among all methods, showing its high computational efficiency.

\begin{figure}
    \centering
    \includegraphics[width=0.45\textwidth]{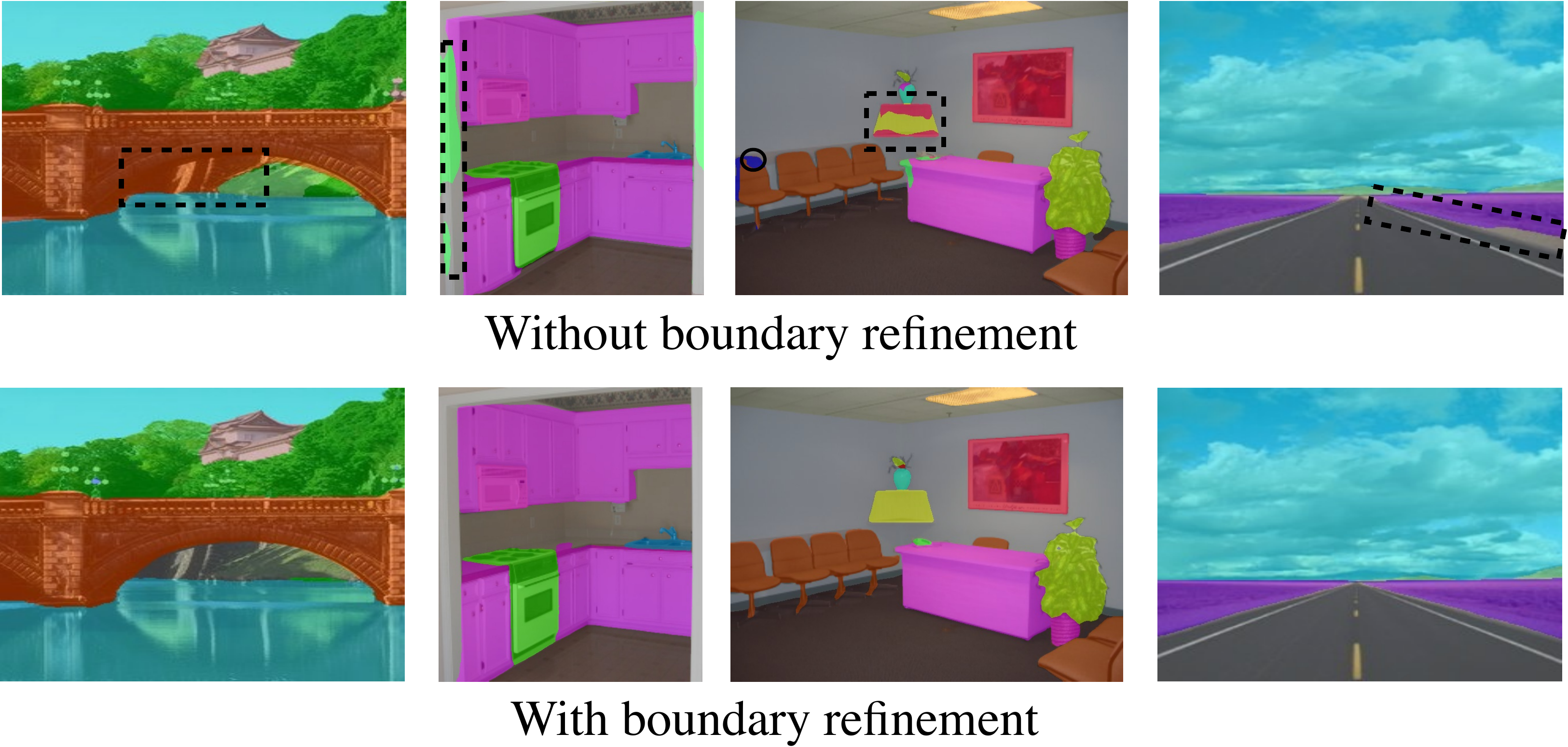}
    \caption{Visualization of segmentation results with/without boundary refinement. Examples are taken from ADE20K validation set.}
    \label{fig:edge_improve}
\end{figure}

\noindent
\textbf{Efficiency of the boundary refinement}
We verify the effectiveness of the spatial branch for boundary refinement in Fig.~\ref{fig:edge_improve}.
Predictions after the refinement module (bottom) often have better boundary estimation than them before the module (top). 

\section{Conclusion}
\label{sec:conclusion}
To summarise, we present a unified framework to tackle the problem of semantic segmentation, by both considering context modeling and boundary refinement.  
We adapt a sparse sampling strategy and use pyramid features to better model contextual information as well as maintaining efficiency.
By adding a spatial path, the model captures dynamic contexts as well as fine-grained boundary signals. 
We hope the proposed UN-EPT method can advocate future work to jointly optimize the contexts and boundary signals for semantic segmentation.


{\small
\bibliographystyle{ieee_fullname}
\bibliography{egbib}
}

\clearpage
\section*{Appendix}
\appendix
\section{Pyramid Features}
\begin{figure}[!htb]
    \centering
    \includegraphics[width=0.45\textwidth]{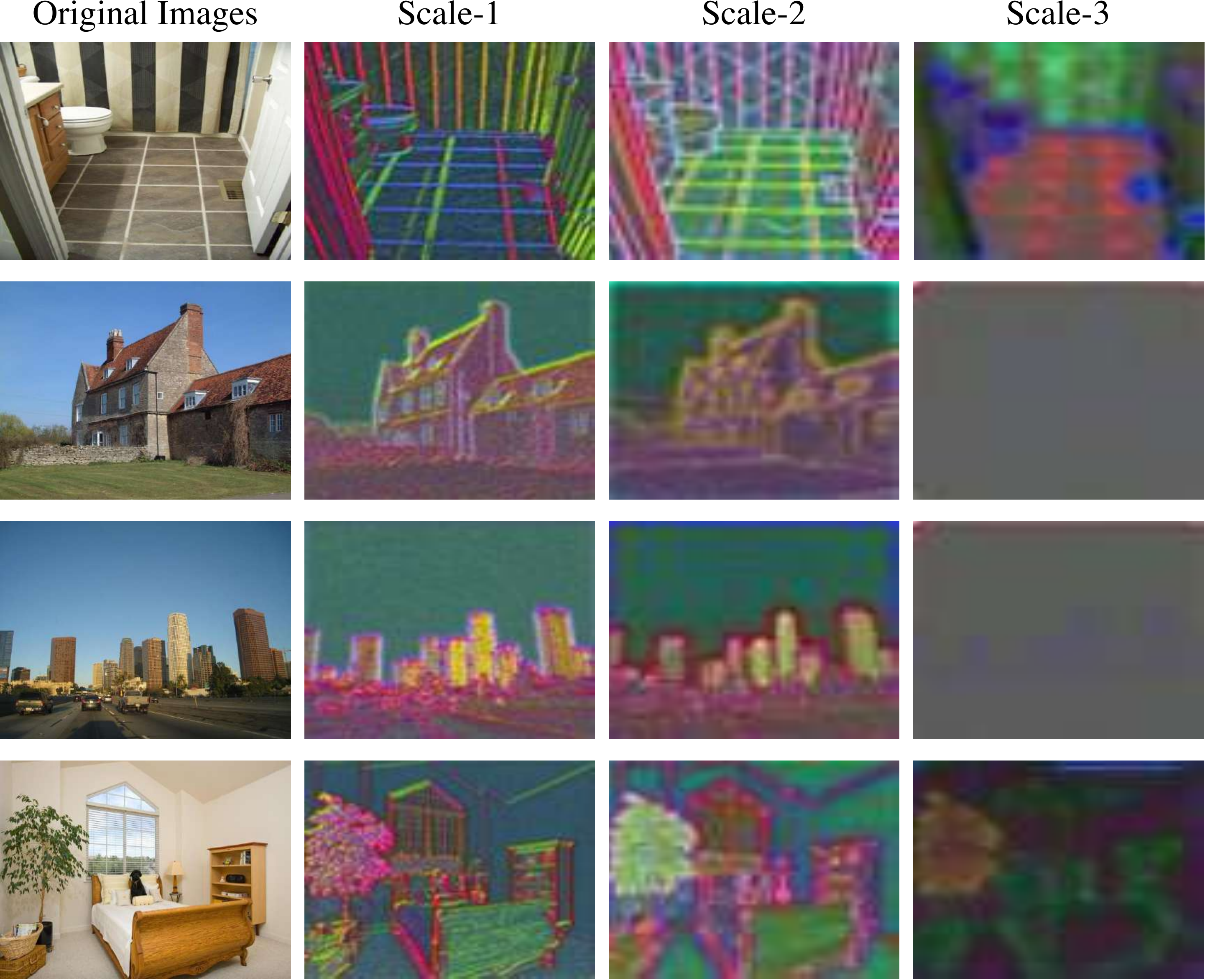}
    \vspace{6pt}
    \caption{Visualization of pyramid feature maps of UN-EPT trained on ADE20K. We take features from C3-C5 stages of the ResNet50 backbone~(\ie,~scale-1 to scale-3). Features from the first scale generally capture more low-level details~(\eg,~boundary statistics), while features from latter layers encode high-level knowledge.
    Best viewed in color.}
    \label{fig:pyramid_feats}
\end{figure}

In the experiments, we adopt three feature maps from different stages of the backbone network. These feature maps are concatenated as a long sequence and serve as the encoder input. We use principal component analysis to reduce the dimension of intermediate feature maps and visualize them in Fig.~\ref{fig:pyramid_feats}. We can see that feature maps from different stages capture image details from different perspectives, providing the transformer encoder with more abundant information.

\section{Backbone Attention}
\begin{figure}[!htb]
    \centering
    \includegraphics[width=0.45\textwidth]{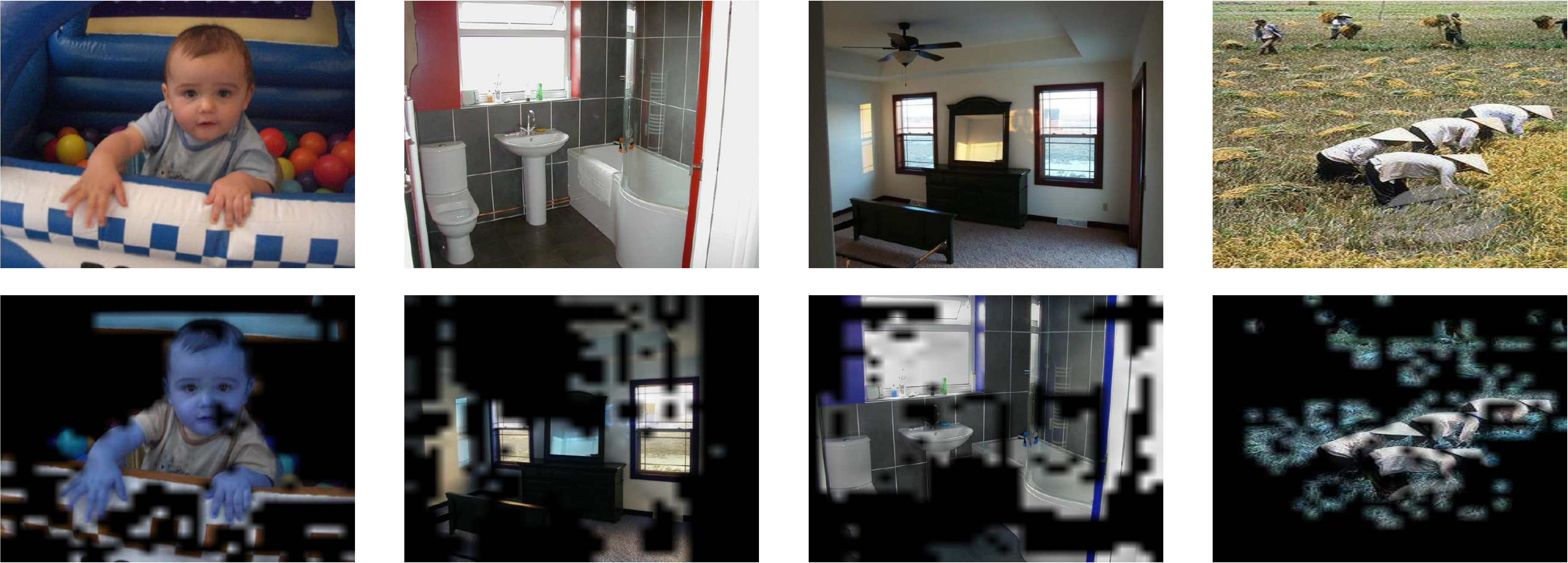}
    \vspace{6pt}
    \caption{Visualization of attention maps of DeiT trained on ADE20K. We compute the joint attention maps from attention weights of 12 layers, where areas with low attention weights are masked. We can see that the backbone can attend to semantically meaningful
    regions, beneficial for semantic segmentation.
    Best viewed in color.}
    \label{fig:deit_attn}
\end{figure}

We use a 12-layer DeiT~\cite{touvron2020training} as our backbone network. We visualize the joint learned attention maps of UN-EPT in Fig.~\ref{fig:deit_attn}. It shows that the backbone network can attend to objects of different scales and is able to capture contextual information.

\section{Sparse Sampling Visualization}
\begin{figure}[!htb]
    \centering
    \includegraphics[width=0.45\textwidth]{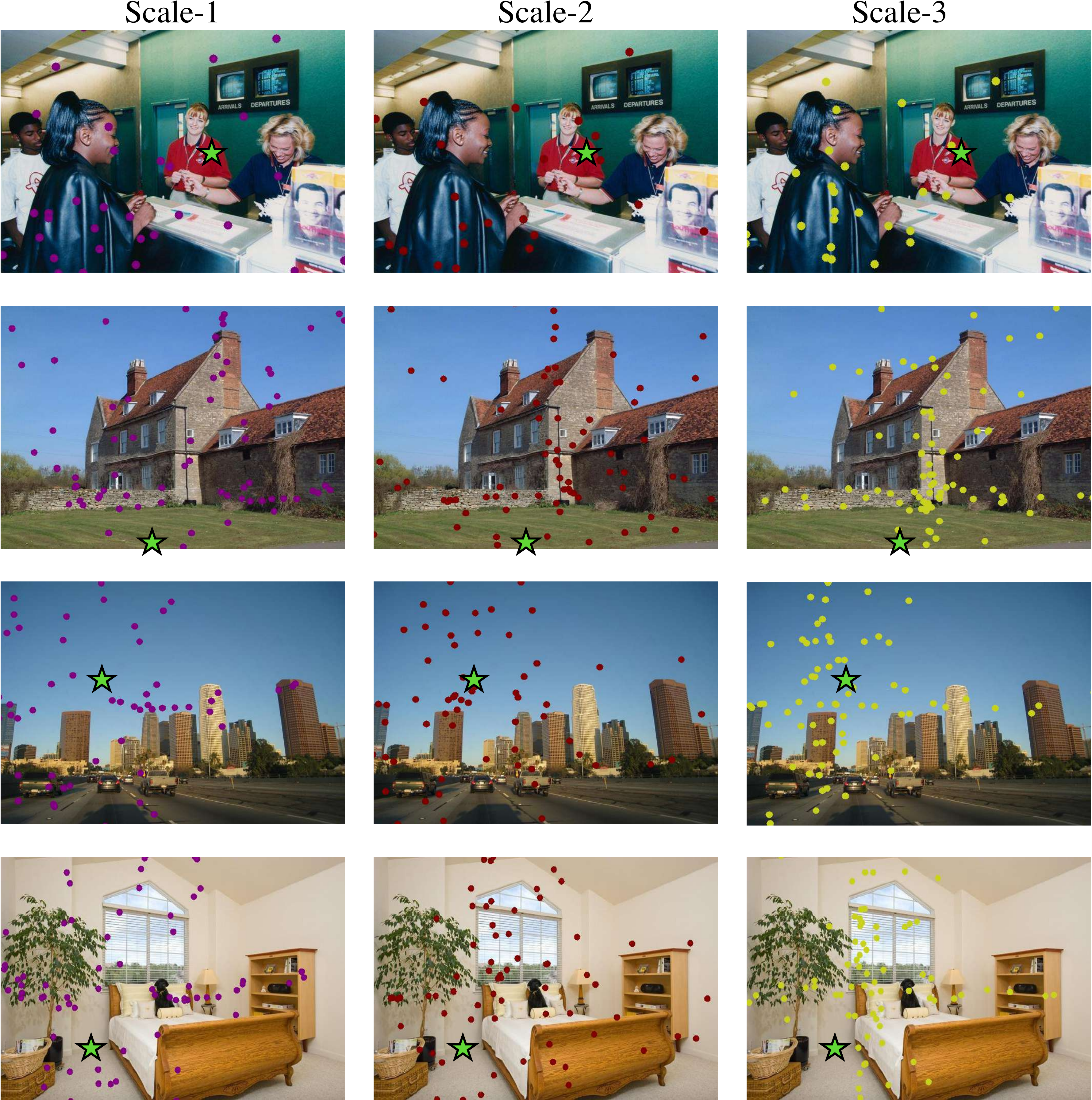}
    \vspace{6pt}
    \caption{Visualization of sampled points of UN-EPT trained on ADE20K. The green five-pointed star denotes a specific query pixel. We show its sampled points on feature maps of three scales.
    For each scale in the figure, a specific query pixel attends to $L \times N \times M $ pixel features~($L$:~scale number, $N$:~\#points per scale, $M$:~head number). We can see that the query pixel attends to different pixels on features of different scales. Compared with full attention, attending to a set of pixels improves the model efficiency a lot. 
    Best viewed in color.}
    \label{fig:sparse_sample}
\end{figure}

We use a sparse sampling strategy to reduce the computational cost of full self-attention. The model learns to only attend to a set of informative pixel features. We show examples in Fig.~\ref{fig:sparse_sample}. From the figure, we can see that a specific query pixel will attend to different sets of points from feature maps of different scales. The distribution of sampled points is able to capture the contextual information of the query pixel. 

\section{Per-class Analysis on ADE20K}
\begin{figure}
    \centering
    \includegraphics[width=0.5\textwidth]{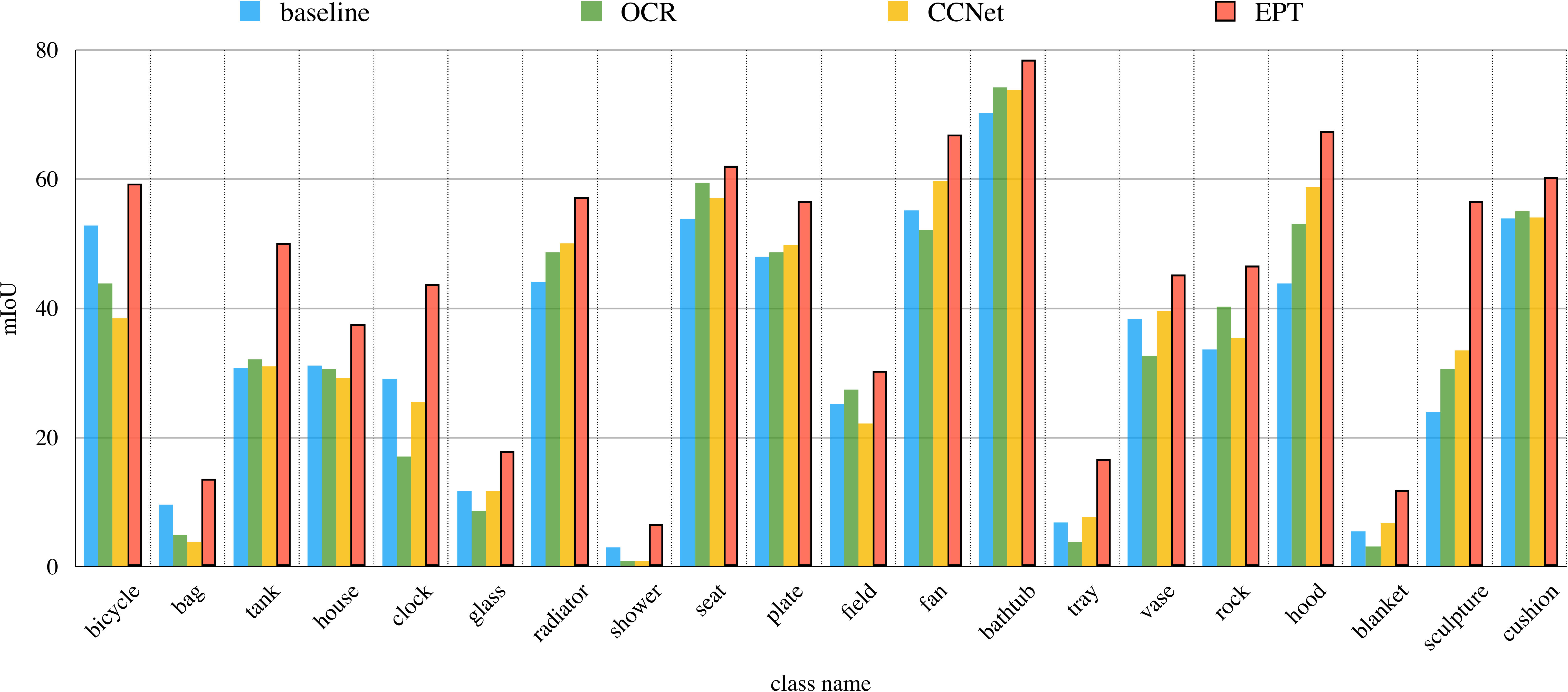}
    \caption{Per-class mIoU results on ADE20K validation set. We choose 20 out of 150 classes for comparison, where UN-EPT outperforms our baseline~(single-scale sparse sampled transformer), CCNet~\cite{huang2019ccnet} and OCRNet~\cite{yuan2019object}. }
    \label{fig:per-class}
\end{figure}

To further analyze the segmentation results and find pros and cons of our method, we choose 20 classes from the total 150 classes in ADE20K, where UN-EPT outperforms state-of-the-art methods, CCNet~\cite{huang2019ccnet} and OCRNet~\cite{yuan2019object}. 
Seen from Fig.~\ref{fig:per-class}, our proposed method is able to better segment objects with complex contexts, \eg~field, rock, hood, which are always surrounded by other candidate objects. Especially, our method shows competitive results in segmenting small objects, \eg~clock, glass, shower, bag, indicating the benefits gained from pyramid features.

\end{document}